\documentclass[10pt,twocolumn,letterpaper]{article}

\usepackage[pagenumbers]{wacv} 

\newcommand{\datasetname}{EvoStruggle}
\newcommand{\papertitle}{EvoStruggle: A Dataset Capturing the Evolution of Struggle across Activities and Skill Levels}
\newcommand{\myparagraph}[1]{\noindent\textbf{#1}}

\definecolor{wacvblue}{rgb}{0.21,0.49,0.74}
\usepackage[pagebackref,breaklinks,colorlinks,allcolors=wacvblue]{hyperref}
\usepackage{multirow}
\usepackage{graphicx}
\usepackage{amsmath}
\usepackage{amssymb}
\usepackage{booktabs}

\def\wacvPaperID{573} 
\def\confName{WACV}
\def\confYear{2026}

\title{\papertitle}

\author{Shijia Feng, Michael Wray, Walterio Mayol-Cuevas\\
University of Bristol\\
{\tt\small shijia.feng.2019, michael.wray, walterio.mayol-cuevas@bristol.ac.uk}
}

\begin{document}
\maketitle
\begin{abstract}
The ability to determine when a person struggles during skill acquisition is crucial for both optimizing human learning and enabling the development of effective assistive systems. As skills develop, the type and frequency of struggles tend to change, and understanding this evolution is key to determining the user’s current stage of learning. However, existing manipulation datasets have not focused on how struggle evolves over time. In this work, we collect a dataset for struggle determination, featuring 61.68 hours of video recordings, 2,793 videos, and 5,385 annotated temporal struggle segments collected from 76 participants. The dataset includes 18 tasks grouped into four diverse activities -- tying knots, origami, tangram puzzles, and shuffling cards, representing different task variations. In addition,  participants repeated the same task five times to capture their evolution of skill. We define the struggle determination problem as a temporal action localization task, focusing on identifying and precisely localizing struggle segments with start and end times. Experimental results show that Temporal Action Localization models can successfully learn to detect struggle cues, even when evaluated on unseen tasks or activities. The models attain an overall average mAP of $34.56\%$ when generalizing across tasks and $19.24\%$ across activities, indicating that struggle is a transferable concept across various skill-based tasks while still posing challenges for further improvement in struggle detection. Our dataset is available at \hyperlink{https://github.com/FELIXFENG2019/EvoStruggle}{https://github.com/FELIXFENG2019/EvoStruggle}
\end{abstract}
\vspace{-5mm}    
\section{Introduction}
\label{sec:intro} 
Understanding human task-completion behaviour requires more than just recognizing success; it also involves analysing the challenges encountered along the way. This highlights the need to model struggle and its evolution as skills develop. Identifying struggling can lead to more effective assistive technologies/teaching systems.

Struggle is characterized by non-smooth, hesitant, repeating, and/or prolonged actions that signal difficulty or uncertainty. It is an inherent part of the human learning process, as people often struggle to develop new skills or complete complex tasks.

Despite its importance, detecting struggle in video remains an under-explored area of research.
However, struggling can be recognised by non-experts and is considered a fundamental aspect of human imitation abilities~\cite{Newell:etal:1991}.
One major challenge is that signs of struggle are often subtle, making them difficult to distinguish from confident actions. Struggling could also manifest differently across different activities. The visual cues used to detect struggle in one domain, such as solving puzzles, may differ from those in another, like assembling mechanical parts, due to variations in movement patterns and task complexity. 
Furthermore, struggle evolves as skill acquisition grows. Capturing this process is important because it can help deal with the nuanced nature of struggle and be a potential marker for which skill stage the performer is at.
These challenges and opportunities highlight the need for dedicated datasets that capture struggle across diverse contexts and stages, enabling the development of robust and generalizable struggle detection models. 

\begin{figure*}[ht]
\begin{center}
\includegraphics[width=1.0\linewidth]{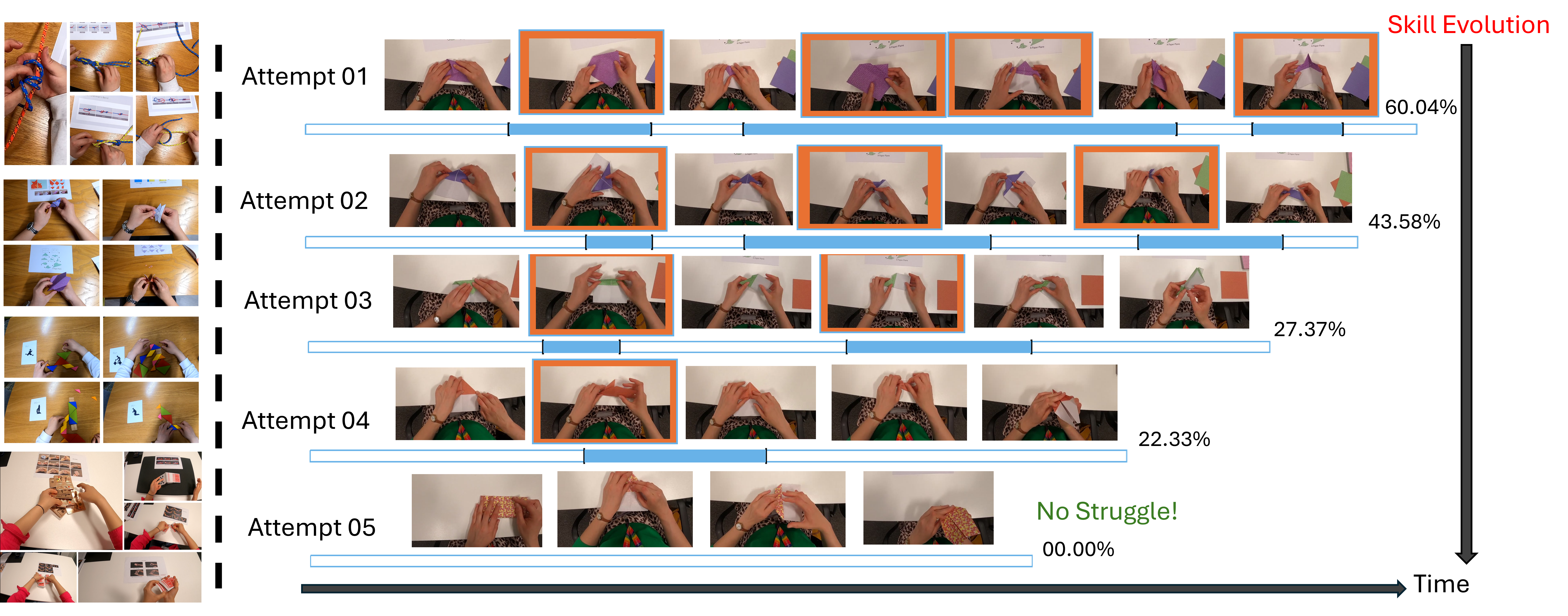}
\end{center}
\vspace{-5mm}
   \caption{Overview of the \datasetname{} dataset. There are four activities: Tying Knots, Origami, Tangrams, and Shuffling Cards, each further consisting of 4/5 distinct tasks (left). Each task has five repetitions that show the evolution of skill (right, top to bottom). Percentages indicate the proportion of struggle duration relative to the total video recording time.}
   \vspace{-3mm}
\label{fig:figure1}
\end{figure*}

In this paper, we introduce \datasetname, a new dataset for struggle determination, comprising over 60 hours of video recordings --  almost $12 \times$ bigger than the previous largest struggle-related dataset~\cite{feng2024strugglingdatasetbaselinesstruggle}.
Participants performed each task five times, demonstrating the evolution of their skill as the proportion of time spent struggling decreased with repeated attempts.
Fig.~\ref{fig:figure1} highlights the dataset's diversity, encompassing 18 tasks grouped into 4 activities, such as folding different origami shapes or shuffling cards in different ways. 
The diversity of our dataset enables a comprehensive evaluation of struggle detection across varied contexts and allows for testing model generalizability across two key gaps: across task and across activity, each with its own challenges.
Another key aspect of \datasetname{} is that the participants repeat their attempts, capturing the evolution of their skill at each task.
This progression adds another crucial dimension along which struggle determination can be understood.

Our contributions can be summarized as follows: 
(i) We present the \datasetname{} dataset, which includes 61.68 hours of video recordings across 18 tasks with 5,385 annotation struggle moments. 
(ii) Our dataset captures participants' repeated attempts at the same task, showcasing the evolution of skill/struggle. 
(iii) We conduct extensive experiments on \datasetname, providing benchmark results for both within activity and across task/activity challenges. 

\section{Related Work}
\label{sec: related work}
\paragraph{Towards Struggle Determination in Video Understanding Datasets.}
Prior research in video understanding has explored various aspects of action recognition and task analysis. These efforts include coarse-grained action recognition\cite{https://doi.org/10.48550/arxiv.1705.06950} and fine-grained action recognition\cite{shao2020finegym, liu2022fineactionfinegrainedvideodataset, https://doi.org/10.48550/arxiv.2204.03646}, as well as workflow analysis in assembly procedures\cite{9022634, sener2022assembly101largescalemultiviewvideo, Ghoddoosian_2023_ICCV}. Other studies\cite{mtlaqa, parmar2019action, Doughty_2017_CVPR, Doughty_2019_CVPR} have aimed to assess task proficiency based on video data. While these approaches provide valuable insights into human actions, they do not explicitly capture struggle---a state characterized by hesitation, failed attempts, and uncertainty.

Struggle determination is a distinct challenge in video understanding, separate from related fields such as skill assessment and error/mistake detection. While skill assessment datasets~\cite{Doughty_2017_CVPR, Doughty_2019_CVPR, grauman2024egoexo4dunderstandingskilledhuman} evaluate proficiency, they do not explicitly measure struggle. Similarly, error/mistake detection datasets~\cite{9022634, sener2022assembly101largescalemultiviewvideo, Ghoddoosian_2023_ICCV, HoloAssist2023, Flaborea_2024_CVPR} focus on identifying mistakes, but struggle does not always equate to making errors—people can struggle without making mistakes and, conversely, can make mistakes without exhibiting signs of struggle.

To address this gap, Feng et al.~\cite{feng2024strugglingdatasetbaselinesstruggle} introduced the first dataset explicitly designed for struggle determination in short video segments. However, struggle determination remains under-explored in large-scale, diverse datasets that span across multiple activities. Our work builds upon this foundation by introducing a significantly larger dataset that captures struggle across numerous participants, various tasks, and repeated attempts, enabling deeper insights into its temporal evolution.

\begin{figure*}[ht]
\begin{center}
\includegraphics[width=\linewidth]{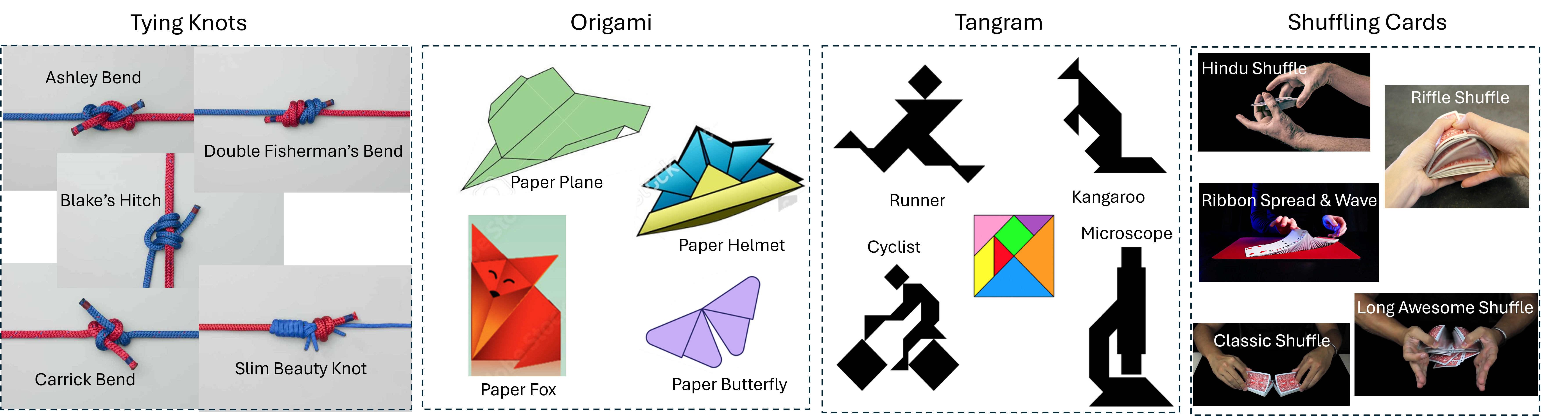}
\end{center}
   \caption{\datasetname{} Dataset Structure. There are four activities: Tying Knots, Origami, Tangram, and Shuffling Cards, each with 4--5 tasks. Each participant completed all tasks from an activity across five attempts to showcase an evolution of skill.}
\label{fig:data-structure}
\vspace{-3mm}
\end{figure*}

\paragraph{Temporal Action Localization for Struggle Action Detection.}  
Our dataset expands on the prior struggle dataset~\cite{feng2024strugglingdatasetbaselinesstruggle} by increasing diversity and improving annotation methods. Unlike prior work that relies on weak labels for short clips, we provide precise temporal boundaries for struggles in untrimmed videos, making Temporal Action Localization (TAL) a natural fit for our task.

TAL focuses on detecting action start and end times within videos and is commonly evaluated using mean Average Precision (mAP) within an Intersection-over-Union (IoU). Prominent TAL benchmark datasets include THUMOS Challenge 2014~\cite{THUMOS14} and ActivityNet-v1.3~\cite{caba2015activitynet}. Compared to Temporal Action Segmentation (TAS), which requires frame-level classification~\cite{farha2019ms, chinayi_ASformer} and post-processing, TAL directly predicts action boundaries, aligning well with our struggle detection goals.

TAL models fall into two categories. Feature-based models, such as AFSD~\cite{lin2021learning}, TadTR~\cite{Liu_2022}, Actionformer~\cite{zhang2022actionformer}, and TriDet~\cite{10203543}, use pre-extracted features from networks like TSN~\cite{wang2016temporal}, I3D~\cite{carreira2017quo}, or SlowFast~\cite{feichtenhofer2019slowfastnetworksvideorecognition}. These models are computationally efficient but rely on pre-trained extractors. Actionformer~\cite{zhang2022actionformer} employs a transformer with a feature pyramid for multi-scale detection, while TriDet~\cite{10203543} introduces Scalable-Granularity Perception (SGP) to reduce self-attention rank loss and a trident head for precise boundary localization.

End-to-end TAL models, such as TALLFormer~\cite{cheng2022tallformertemporalactionlocalization}, Re2TAL~\cite{Zhao_2023_CVPR}, and ViT-TAD~\cite{yang2024adapting}, train feature extraction and detection jointly but require significant GPU memory. To mitigate this, TALLFormer~\cite{cheng2022tallformertemporalactionlocalization} updates only one snippet per iteration, ViT-TAD~\cite{yang2024adapting} extends transformers for long-term sequences, and Re2TAL~\cite{Zhao_2023_CVPR} introduces reversible modules to free memory caches.

Our dataset's focus on struggle detection in untrimmed videos necessitates precise temporal annotations, making TAL methods a suitable baseline.

\section{\datasetname{} Dataset}
\label{sec:dataset}
In this section, we introduce \datasetname, describe its collection and annotation process and present key statistics.

\subsection{Dataset Overview} 
\label{data overview} 
Inspired by the definition of struggle in~\cite{feng2024strugglingdatasetbaselinesstruggle}, we define struggle as \emph{``Observable difficulty towards completing a given activity''}. This could be represented by motor hesitation of hands, repeated attempts, prolonged actions, signs of frustration (e.g. through hand and or head movements), or disruptive errors and pauses.
Note that these can be task-specific: signs of struggle for one activity could be normal operations for a separate activity. 
For example, repeated attempts often signal struggle in tasks like knot tying, origami, and tangram, where participants retry actions when stuck at certain stages. In contrast, repeatedly shuffling cards is typically not a sign of struggle---this reflects common card game behaviour.

We chose activities based on three principles: activities should be accessible to participants, each activity can have many separate related tasks, and participants will struggle if they are not experts or familiar with the activity/task.
Following these criteria, we select Tying Knots, Origami, Tangram, and Shuffling Cards as our activities to match these constraints.
These activities require careful manipulative motions (Origami/Tying Knots), trial-and-error placement and visual search (Tangram), and fast-paced dexterity (Shuffling Cards), covering diverse types of struggle. While they share desk-based setups, they differ in materials, room settings, and visual appearances, and task-level variations introduce further behavioural and manipulation diversity.

To investigate how struggle evolves with practice/experience, we asked participants to repeat each task five times. This setting was chosen based on prior experiments, balancing the need to capture learning progression without causing participant fatigue. This repetition allowed us to observe how struggle moments changed with increasing familiarity and skill.

The overall structure of \datasetname \space is illustrated in Fig.~\ref{fig:data-structure}. It is organized into four activities: Tying Knots, Origami, Tangram, and Shuffling Cards. Each activity includes multiple participants, with every participant completing several tasks and repeating each task five times.

\begin{figure*}[ht]
\begin{center}
\includegraphics[width=\linewidth]{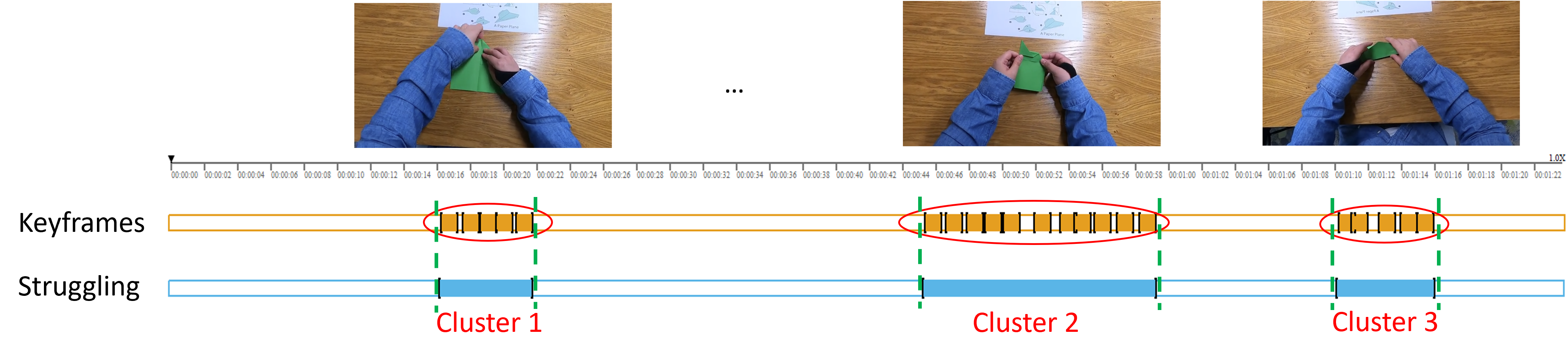}
\end{center}
\vspace{-5mm}
   \caption{Our Struggle Annotation Pipeline consists of two stages. First, annotators watch the video and indicate moments whenever they believe the person is struggling. In the second stage, we cluster these moments into contiguous start/end times.}
\label{fig:annotation-pipeline}
\end{figure*}

\begin{table*}[t]
\scriptsize
\begin{center}
\begin{tabular}{lrrrrrrrr}
\toprule
 & \multicolumn{5}{c}{Total} & \multicolumn{3}{c}{Per Video} \\ \cmidrule(lr){2-6} \cmidrule(lr){7-9}
Activities & Participants & Videos & Struggle Inst. & Struggle Dur. (hrs) & Rec. Dur. (hrs) & Struggle Inst. & Struggle Dur. (s) & Rec. Dur. (s) \\ \midrule
Tying Knots & 34 & 806 & 1167 & 6.85 & 13.44 & 1.45$\pm$1.20 & 30.61$\pm$36.18 & 60.05$\pm$45.59 \\
Origami & 32 & 637 & 974 & 6.54 & 17.32 & 1.53$\pm$1.57 & 36.94$\pm$58.22 & 97.86$\pm$60.79 \\
Tangram & 30 & 600 & 1098 & 9.73 & 14.40 & 1.83$\pm$1.30 & 58.38$\pm$59.79 & 86.39$\pm$63.22 \\
Shuffle Cards & 30 & 750 & 2146 & 4.98 & 16.52 & 2.86$\pm$1.89 & 23.90$\pm$19.45 & 79.31$\pm$17.66 \\ \midrule
\textbf{Total} & \textbf{126 (76 unique)} & \textbf{2793} & \textbf{5385} & \textbf{28.10} & \textbf{61.68} & \textbf{1.93$\pm$1.62} & \textbf{36.22$\pm$46.62} & \textbf{79.50$\pm$50.78} \\ \bottomrule
\end{tabular}%
\end{center}
\vspace{-5mm}
\caption{\datasetname{} Statistics. For each activity, we show the number of participants and the number of recorded videos. Additionally, we provide the number of struggle instances, struggle duration, and recording duration overall and per video (mean$\pm$std). Comparison with the existing struggle dataset~\cite{feng2024strugglingdatasetbaselinesstruggle} is provided in the supplementary material.}
\vspace{-3mm}
\label{tab:overall-statistics}
\end{table*}

\subsection{Dataset Collection} 
\label{data collection} 

We recruited participants and prioritised those with no experience in the activity.
Videos were captured using head-mounted GoPro Hero 8 cameras.
The cameras were recorded at a resolution of $1920\times1080$ with a 50 FPS frame rate, using a standard or wide field of view to ensure that participants’ hands, objects, and printed instructions remained fully visible throughout the activity. 

Each session lasted approximately 30 minutes, during which participants completed all tasks within an activity according to the provided instructions, repeating each task five times.
We used paper-printed instructions, which included only key steps or final goal patterns, placing them in front of participants during video recording.
This approach ensured that participants had the necessary information while maintaining a level of challenge. 
In the Tangram activity, if a participant didn't finish the task after 3 minutes, the attempt was stopped, and some hints were given to them before their next attempt. We found that struggling beyond 3 minutes almost entirely repeated previous patterns (i.e., pausing).

Participants could complete multiple activities, but were restricted from repeating the same activity to maintain data diversity and avoid potential data leakage when the data was divided for training. In total, we had 76 unique participants, where 46 took part in only one activity, 19 participants took part in two activities, 7 participants took part in three activities, and 4 participants took part in all four activities.

\subsection{Annotating Struggle}
\label{annotation}

We annotate the start and end time boundaries to capture the moments when participants struggle. 
Our struggle annotation approach is illustrated in Fig.~\ref{fig:annotation-pipeline}, which outlines our two-stage approach.
In the first stage, we identify ‘keyframes’ by reviewing the video and marking moments where the person appears to struggle. Once a struggle moment is detected, we continue monitoring to capture additional keyframes.
The keyframes naturally form clusters, and so, in the second stage, we define the start and end boundaries of a struggle instance by taking the leftmost and rightmost keyframes within each cluster.
We found this approach to result in high-quality annotations at a $2\times$ speed-up over manually annotating start/end times.
We annotated our videos using one expert annotator to keep consistency, since our pilot study involving non-experts led to inconsistent and noisy annotations.
A bowser-based video annotation software, VIA Video Annotator~\cite{dutta2019vgg}, was used for annotating struggle. 

\begin{figure*}[ht]
\begin{center}
\includegraphics[width=\linewidth]{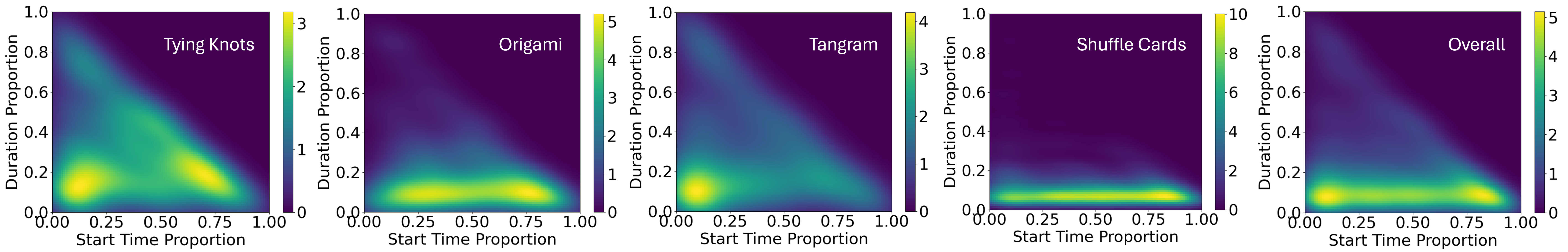}
\end{center}
\vspace{-5mm}
   \caption{Heatmaps showing the Kernel Density Estimation (KDE) of struggle instance distributions. The x-axis/y-axis represents the normalized start/duration time of struggle relative to the total video recording time.}
   \vspace{-3mm}
\label{fig:heatmap-struggle-start-duration-kde}
\end{figure*}

\subsection{\datasetname{} Statistics}
\label{dataset statistics}

\begin{figure}[ht]
\centering
\includegraphics[width=\linewidth]{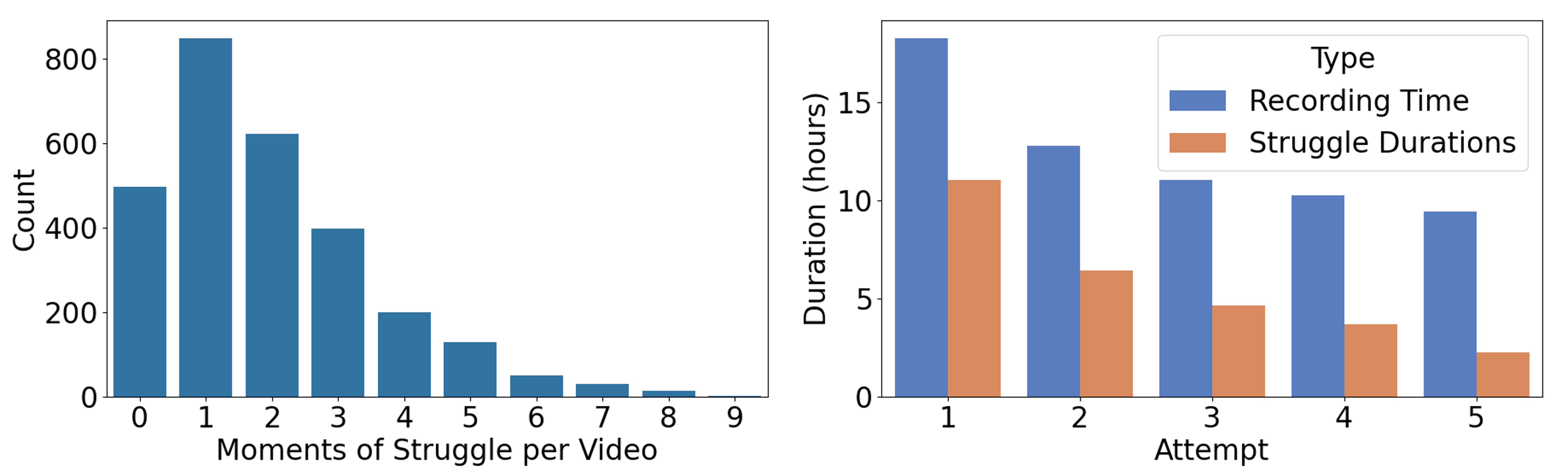}
\caption{Number of struggle moments per video (left) and total recording time with struggle durations per attempt (right) in \datasetname.}
\vspace{-3mm}
\label{fig:combined_struggle_figures}
\end{figure}

The statistics of our dataset are summarized in Table~\ref{tab:overall-statistics}.
A total of 76 participants contributed to 126 video recording sessions, with each activity involving at least 30 participants.
The dataset comprises of 2,793 videos containing 5,385 annotated temporal struggle instances. The dataset has 61.68 hours of recording, of which 28.1 hours are labelled as the participant struggling.
The average struggle duration varies across activities, ranging from 23.90 seconds in the card-shuffling activity to 58.38 seconds in the tangram activity. Similarly, the average video duration ranges from 60.05 seconds for tying knots to 97.86 seconds for origami. 

We show the number of instances of struggle per video on the left side of Fig.~\ref{fig:combined_struggle_figures}, showing participants generally struggled less than 5 times per video, though it exhibits a long-tail-like distribution with up to 9 unique struggle moments in a video. 
We also showcase recording time and struggle duration per attempt on the right side of Fig.~\ref{fig:combined_struggle_figures}.
The figure highlights that early attempts have a high percentage of struggle moments ($60\%$) and take longer, whereas later attempts have a much lower percentage of struggle moments ($24\%$) and are shorter as participants' skill at the activity improves.

Finally, we visualize the annotated struggle moment distributions in Fig.~\ref{fig:heatmap-struggle-start-duration-kde}. These heatmaps show how struggle moments are distributed across videos, helping to identify potential biases in model training.
The tying knots activity has the most diverse struggle distribution, spanning the full range of start times and durations. Origami and card-shuffling also show diverse start times but with struggle durations mostly within the first 20\% of the normalized duration. In tangram, struggle moments are concentrated in the first 20\% of the timeline, likely due to initial confusion in the puzzle-solving task.
Overall, struggle moments tend to cluster at the beginning (around 20\%) and end (around 80\%) of activities, though they can occur at any point.

\section{Experiments}
\label{sec: experiments} 
In this section, we provide baseline results for \datasetname, and structure experiments to answer the following questions: (i) Can action localization models localize struggle moments effectively? (ii) How generalizable are the models from a task and activity perspective? (iii) How does the evolution of skill change struggle localization performance? and (iv) What qualitative analysis can be performed?

\myparagraph{Evaluation Metrics}
We report struggle temporal localization evaluation results in mean Average Precision (mAP) over different threshold Intersection-over-Unions (tIoU) ($\{0.3, 0.5, 0.7\}$), as well as the averaged mAP over the different thresholds.
Further thresholds are presented in supp. 

\myparagraph{Baseline Models}
We choose two feature-based TAL models, Actionformer~\cite{zhang2022actionformer} and TriDet~\cite{10203543}, in addition to the end-to-end TAL model Re2TAL~\cite{Zhao_2023_CVPR}, which utilizes reversible SlowFast-101~\cite{feichtenhofer2019slowfastnetworksvideorecognition} as the backbone and Actionformer~\cite{zhang2022actionformer} as the action detection head, to act as baselines for \datasetname. These models represent recent SOTA methods for the Temporal Action Localization Task. 
Full details of these models can be found in the supp.

\begin{figure}[t]
\begin{center}
\includegraphics[width=\linewidth]{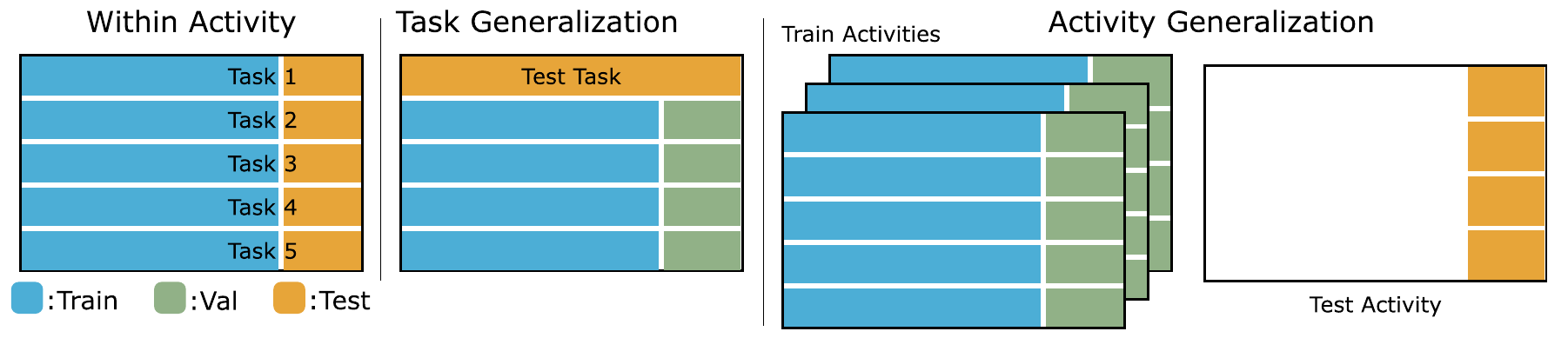}
\end{center}
\vspace{-3mm}
   \caption{Train/Val/Test splits for Within Activity, Task Generalization, and Activity Generalization tasks. Note that for Activity Generalization, the test set consists of the validation sets from the withheld activity.}
   \vspace{-3mm}
\label{fig:data-splits}
\end{figure}

\myparagraph{Dataset Splits}
We define three separate splits, which can be seen in Figure~\ref{fig:data-splits}, depending on the level of generalization we wish to test across the different models.
\textbf{Within Activity:} An activity is split by a ratio of 7:3/8:2 (based on the number of tasks) such that all tasks are seen in training, and no participant is seen in both train and test sets.
\textbf{Task Generalization:} We perform leave one out cross-fold validation across the tasks within a single activity. A small subset of the tasks in training is withheld to use as a validation set.
\textbf{Activity Generalization:} We perform leave one out cross-fold validation across the activities. A subset of the activities for training is withheld as a validation set.

\subsection{Within-Activity Evaluation}
\label{within-domain-eval}

\begin{table}[ht]
\centering
\resizebox{0.45\textwidth}{!}{%
\begin{tabular}{ll|cccc}
\toprule
\multirow{2}{*}{Activity} & \multirow{2}{*}{Model} & \multicolumn{4}{c}{mAP@tIoU and Average} \\ \cmidrule{3-6}
 &  & 0.3 & 0.5 & 0.7 & Avg. \\ \midrule
\multirow{4}{*}{Tying Knots} & Random & 8.80\% & 1.79\% & 0.19\% & 3.17\% \\
 & Actionformer~\cite{zhang2022actionformer} & 67.99\% & 39.21\% & 10.38\% & 39.39\% \\
 & TriDet~\cite{10203543} & 65.44\% & 43.91\% & 14.53\% & \textbf{41.93\%} \\
 & Re2TAL~\cite{Zhao_2023_CVPR} & 61.73\% & 38.12\% & 8.70\% & 35.92\% \\ \midrule
\multirow{4}{*}{Origami} & Random & 7.25\% & 0.97\%  & 0.06\% & 2.42\% \\
 & Actionformer~\cite{zhang2022actionformer} & 52.75\% & 27.73\% & 5.23\% & 27.98\% \\
 & TriDet~\cite{10203543} & 54.32\% & 27.00\% & 5.73\% & 28.38\% \\
 & Re2TAL~\cite{Zhao_2023_CVPR} & 57.37\% & 29.62\% & 8.78\% & \textbf{32.36\%} \\ \midrule
\multirow{4}{*}{Tangram} & Random & 10.29\% & 1.90\% & 0.20\% & 3.46\% \\
 & Actionformer~\cite{zhang2022actionformer} & 55.85\% & 30.64\% & 4.90\% & 29.95\% \\
 & TriDet~\cite{10203543} & 57.21\% & 29.77\% & 6.21\% & 30.70\% \\
 & Re2TAL~\cite{Zhao_2023_CVPR} & 69.27\% & 44.83\% & 19.11\% & \textbf{44.57\%} \\ \midrule
\multirow{4}{*}{Shuffle Cards} & Random & 5.07\% & 0.84\% & 0.08\% & 1.66\% \\
 & Actionformer~\cite{zhang2022actionformer} & 71.49\% & 56.40\% & 21.85\% & 50.80\% \\
 & TriDet~\cite{10203543} & 70.94\% & 55.26\% & 20.28\% & 49.55\% \\
 & Re2TAL~\cite{Zhao_2023_CVPR} & 78.26\% & 62.30\% & 35.21\% & \textbf{59.77\%} \\ \bottomrule \\
\end{tabular}%
}
\vspace{-3mm}
\caption{Within-Activity Evaluation Experiment Results. The results are reported on the validation set in each activity.}
\vspace{-3mm}
\label{tab:within-domain-eval}
\end{table}

We first benchmark methods on their ability to localize struggle within activities in Table~\ref{tab:within-domain-eval}.
We show that the average mAP across activities ranges from 27.98\% to 59.77\%, with the highest overall performance across models on the shuffling cards activity. This suggests that struggle in this activity has more distinct patterns of struggle, such as dropping cards. Models' performance on Tying Knots and Tangram exhibit intermediate performance, while Origami appears to be the most challenging activity for struggle localization.
This is likely due to the subtle and fine-grained nature of struggle in origami, where difficulties may manifest as slight hesitations, slower movements, or repeated attempts at certain steps. Additionally, Tangram shows a larger performance gap between models, up to 14.62\%, suggesting greater variability in struggle patterns within this activity.

Among the evaluated models, Re2TAL~\cite{Zhao_2023_CVPR} achieves the highest average mAP in three out of four activities, demonstrating its effectiveness in localizing struggle instances. This advantage is likely due to its end-to-end training approach, which contrasts with the feature-based methods Actionformer~\cite{zhang2022actionformer} and TriDet~\cite{10203543}. Notably, Re2TAL~\cite{Zhao_2023_CVPR} performs particularly well in tangram and shuffle cards, where it significantly outperforms the other models by a large margin (up to 14.62\% in tangram and 10.22\% in shuffle cards).
However, it falls behind TriDet~\cite{10203543} in tying knots. This could be due to the inherent complexity of the tying knots task, which involves intricate hand movements and varying struggle durations. As shown in the heatmaps in Fig.~\ref{fig:heatmap-struggle-start-duration-kde}, struggle instances in tying knots are widely distributed across both start times and normalized durations. This variability may make Re2TAL~\cite{Zhao_2023_CVPR} more prone to focusing on a specific range of struggle start times or durations during end-to-end training, potentially limiting its generalization. These baseline results provide insights into current model performance and showcase a large gap, especially at higher IoUs, for future work to investigate.

\subsection{Struggle Generalization}
In this section, we describe the experimental procedures for evaluating task-level and activity-level generalization, along with the corresponding results and discussions.
\textbf{Task-level generalization} evaluates the model's ability to generalize across tasks within the same activity, i.e. testing on unseen tasks from the same activity.
This assesses whether the model can effectively capture shared features within an activity. 
\textbf{Activity-level generalization} examines the model's ability to generalize across different activities by testing on unseen activities and evaluating its adaptability to distinct task categories.
\vspace{-2mm}
\paragraph{Task-level Generalization}
We aim to address whether common features can be shared to detect struggle across various skill-performing scenarios. Ideally, the visual features used for determining struggle should not be domain-specific, but rather generalizable across domains with similar actions, such as peeling an onion and peeling an apple. Leveraging the diversity of multiple activities and tasks within the activities in our new struggle dataset, we evaluate the models' generalizability in detecting struggle moments.

As shown in Fig.~\ref{fig:data-splits}, for each activity, we hold out one task at a time as the test set to evaluate the `cross task' performance of the temporal struggle action localization while we train the deep models on a combination of the rest of the tasks within the same activity domain. 

\begin{table}[ht]
\centering
\resizebox{0.45\textwidth}{!}{%
\begin{tabular}{ll|cccccc}
\toprule
\multirow{2}{*}{Activity} & \multirow{2}{*}{Model} & \multicolumn{5}{c}{Average mAP@tIoU} \\ \cmidrule{3-8}
 & & Task 01 & Task 02 & Task 03 & Task 04 & Task 05 & Average \\
\midrule
\multirow{4}{*}{Tying Knots} & Random & 5.59\% & 5.39\% & 8.17\% & 5.90\% & 2.94\% & 5.60\% \\
 & Actionformer~\cite{zhang2022actionformer} & 38.21\% & 36.63\% & 36.42\% & 29.67\% & \textbf{27.58\%} & 33.70\% \\
 & TriDet~\cite{10203543} & \textbf{43.70\%} & \textbf{43.31\%} & 42.43\% & \textbf{30.79\%} & 24.71\% & \textbf{36.99\%} \\
 & Re2TAL~\cite{Zhao_2023_CVPR} & 40.54\% & 40.36\% & \textbf{46.96\%} & 28.11\% & 20.47\% & 35.29\% \\
\midrule
\multirow{4}{*}{Origami} & Random & 3.72\% & 3.53\% & 2.77\% & 2.91\% & - & 3.23\% \\
 & Actionformer~\cite{zhang2022actionformer} & 24.69\% & 18.22\% & \textbf{23.03\%} & 23.40\% & - & 22.34\% \\
 & TriDet~\cite{10203543} & 23.65\% & 21.03\% & 20.70\% & 21.59\% & - & 21.74\% \\
 & Re2TAL~\cite{Zhao_2023_CVPR} & \textbf{34.92\%} & \textbf{25.03\%} & \textbf{23.05\%} & \textbf{26.77\%} & - & \textbf{27.44\%} \\
\midrule
\multirow{4}{*}{Tangram} & Random & 6.80\% & 5.42\% & 4.17\% & 4.66\% & - & 5.26\% \\
 & Actionformer~\cite{zhang2022actionformer} & 29.63\% & 28.37\% & 20.59\% & 33.90\% & - & 28.12\% \\
 & TriDet~\cite{10203543} & 30.50\% & 32.02\% & 23.27\% & 34.08\% & - & 29.97\% \\
 & Re2TAL~\cite{Zhao_2023_CVPR} & \textbf{33.38\%} & \textbf{43.50\%} & \textbf{34.11\%} & \textbf{45.97\%} & - & \textbf{39.24\%} \\
\midrule
\multirow{4}{*}{Shuffle Cards} & Random & 1.63\% & 2.28\% & 1.92\% & 2.54\% & 2.08\% & 2.09\% \\
 & Actionformer~\cite{zhang2022actionformer} & 11.48\% & 29.31\% & 31.55\% & 34.21\% & 33.05\% & 27.92\% \\
 & TriDet~\cite{10203543} & 9.70\% & 32.12\% & 27.29\% & 32.01\% & 34.12\% & 27.05\% \\
 & Re2TAL~\cite{Zhao_2023_CVPR} & \textbf{15.10\%} & \textbf{38.56\%} & \textbf{33.26\%} & \textbf{36.30\%} & \textbf{49.71\%} & \textbf{34.59\%} \\
\bottomrule \\
\end{tabular}
}%
\vspace{-3mm}
\caption{Task Generalization Experiment Results. The results are reported using averaged mAPs where the models are evaluated on the held-out task. The rightmost column is the average of mAP performance over all the tasks within each of the activities.}
\vspace{-3mm}
\label{tab:task-level gen}
\end{table}

The task-level generalization results are presented in Table~\ref{tab:task-level gen}. The averaged mAPs are computed for each hold-out task using various models across all four activities in our dataset, with the overall average across tasks included in the last column of the table. As a baseline, we also provide results for random performance. This baseline uses the frequency distribution of struggle segments in the training set as a probability distribution to generate a certain number of struggle segments during evaluation, assigning random start and end times and calculating mAPs accordingly and serving as a reference point. The table shows that the task-level generalization results significantly outperform the random baseline, with the overall averaged mAPs ranging from 20\% to 40\%, compared to the random baseline's range of 2\% to 5.60\%. These findings suggest that the deep model parameters trained on a variety of tasks are effective for detecting struggle in unseen tasks within the same activity domain. This indicates that the features learned by the models for struggle detection share commonalities across different tasks.

In terms of model performance, the two feature-based models, Actionformer~\cite{zhang2022actionformer} and TriDet~\cite{10203543} achieve comparable averaged mAPs. However, the end-to-end model, Re2TAL~\cite{Zhao_2023_CVPR}, generally achieves significantly higher mAPs, except for the tying knots activity, where its average performance across tasks is 35.29\%, compared to the highest performance of 36.99\%. We attribute the superior performance of Re2TAL~\cite{Zhao_2023_CVPR} to the joint training of the feature extraction backbone, as it further fine-tuned the backbone parameters during the training stage to better extract useful spatial-temporal features for struggle detection. 

\paragraph{Activity Generalization}
\vspace{-5mm}
\begin{table}[ht]
\centering
\resizebox{0.45\textwidth}{!}{%
\begin{tabular}{ll|cccc}
\toprule
\multirow{2}{*}{Activity} & \multirow{2}{*}{Model} & \multicolumn{4}{c}{mAP@tIoU and Average} \\ \cmidrule{3-6}
 & & 0.3 & 0.5 & 0.7 & Avg. \\ \midrule
\multirow{4}{*}{Tying Knots} & Random & 8.80\% & 1.79\% & 0.19\% & 3.17\% \\
 & Actionformer~\cite{zhang2022actionformer} & 45.13\% & 20.37\% & 4.30\% & 22.58\% \\
 & TriDet~\cite{10203543} & 34.76\% & 14.08\% & 2.68\% & 16.25\% \\
 & Re2TAL~\cite{Zhao_2023_CVPR} & 47.14\% & 23.45\% & 5.19\% & \textbf{25.05\%} \\ \midrule
\multirow{4}{*}{Origami} & Random & 7.25\% & 0.97\% & 0.06\% & 2.42\% \\
 & Actionformer~\cite{zhang2022actionformer} & 32.07\% & 8.54\% & 0.99\% & \textbf{12.24\%} \\
 & TriDet~\cite{10203543} & 28.78\% & 7.08\% & 1.12\% & 10.72\% \\
 & Re2TAL~\cite{Zhao_2023_CVPR} & 26.26\% & 9.14\% & 2.65\% & 11.67\% \\ \midrule
\multirow{4}{*}{Tangram} & Random & 10.29\% & 1.90\% & 0.20\% & 3.46\% \\
 & Actionformer~\cite{zhang2022actionformer} & 44.58\% & 15.60\% & 2.83\% & 19.60\% \\
 & TriDet~\cite{10203543} & 47.07\% & 18.19\% & 3.23\% & 21.42\% \\
 & Re2TAL~\cite{Zhao_2023_CVPR} & 49.53\% & 24.12\% & 8.97\% & \textbf{26.98\%} \\ \midrule
\multirow{4}{*}{Shuffle Cards} & Random & 5.07\% & 0.84\% & 0.08\% & 1.66\% \\
 & Actionformer~\cite{zhang2022actionformer} & 29.15\% & 9.86\% & 1.40\% & \textbf{12.69\%} \\
 & TriDet~\cite{10203543} & 28.42\% & 7.15\% & 0.67\% & 10.75\% \\
 & Re2TAL~\cite{Zhao_2023_CVPR} & 24.80\% & 8.00\% & 0.98\% & 10.53\% \\ \bottomrule \\
\end{tabular}%
}
\vspace{-3mm}
\caption{Activity-Level Generalization Experiment Results. The results are reported based on the validation set in each activity as the held-out test activity.}
\label{tab:activity-level gen}
\end{table}
\vspace{-5mm}
\begin{figure}[ht]
\begin{center}
\includegraphics[width=0.7\linewidth]{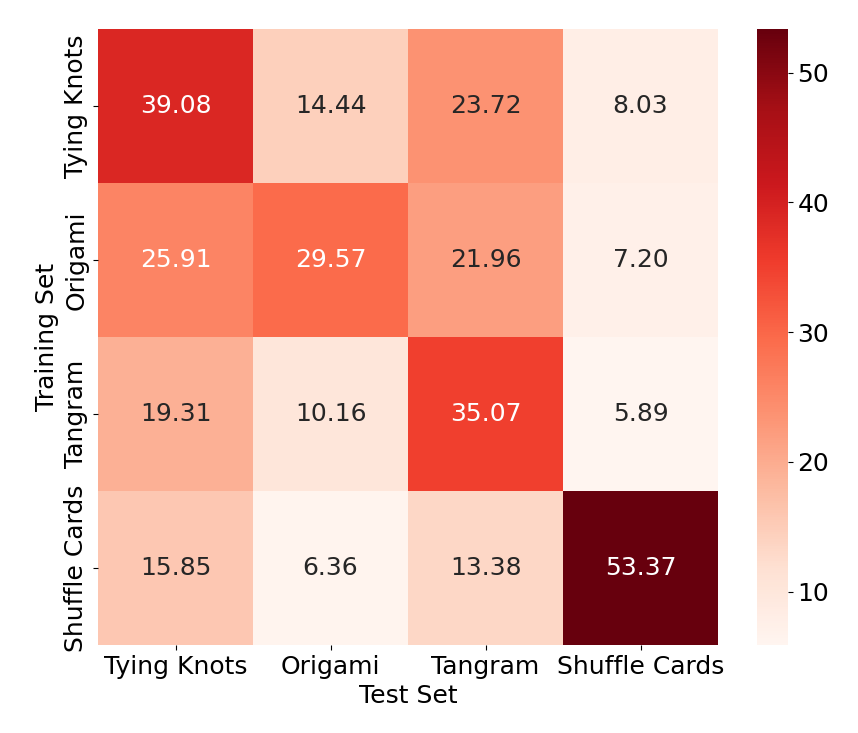}
\end{center}
\vspace{-5mm}
   \caption{Heatmaps showing the alternate training and evaluation performance of activity-level generalization.}
\label{fig:activity-level-gen-matrix-heatmap}
\end{figure}

To evaluate activity generalization, we hold out videos from the validation set of one activity at a time as the test set and combine the train and validation (val) sets of the other three activities for training (see Fig.~\ref{fig:data-splits}).

The main activity-level generalization results are shown in Table~\ref{tab:activity-level gen}.
Naturally, this is a challenging task as the four activities are quite different from one another.
However, we still see a clear improvement over random, proving that some characteristics of struggle determination are universal.
Notably, Re2TAL~\cite{Zhao_2023_CVPR} outperforms other methods on the Tying Knots and Tangram tasks but struggles compared to Actionformer~\cite{zhang2022actionformer} on both Origami and Shuffling Cards.
This could be due to the end-to-end model overfitting slightly during training and struggles to generalize across activities. 

Next, we explore generalization across activities individually, with all combinations of activities used as train and test, Fig.~\ref{fig:activity-level-gen-matrix-heatmap} presents these results.
Interestingly, shuffling cards is the least helpful as a training activity and the hardest activity to generalize to, whereas the results suggest that the other three activities share a greater overlap.
We believe this is because the visual cues for detecting struggle in the shuffling cards activity are distinct from those in the other three. In particular, people may struggle when dropping a lot of cards, a scenario that does not occur in the other activities. 

\begin{figure}[ht]
\centering
\begin{subfigure}{1.\linewidth}
    \centering
    \includegraphics[width=\linewidth]{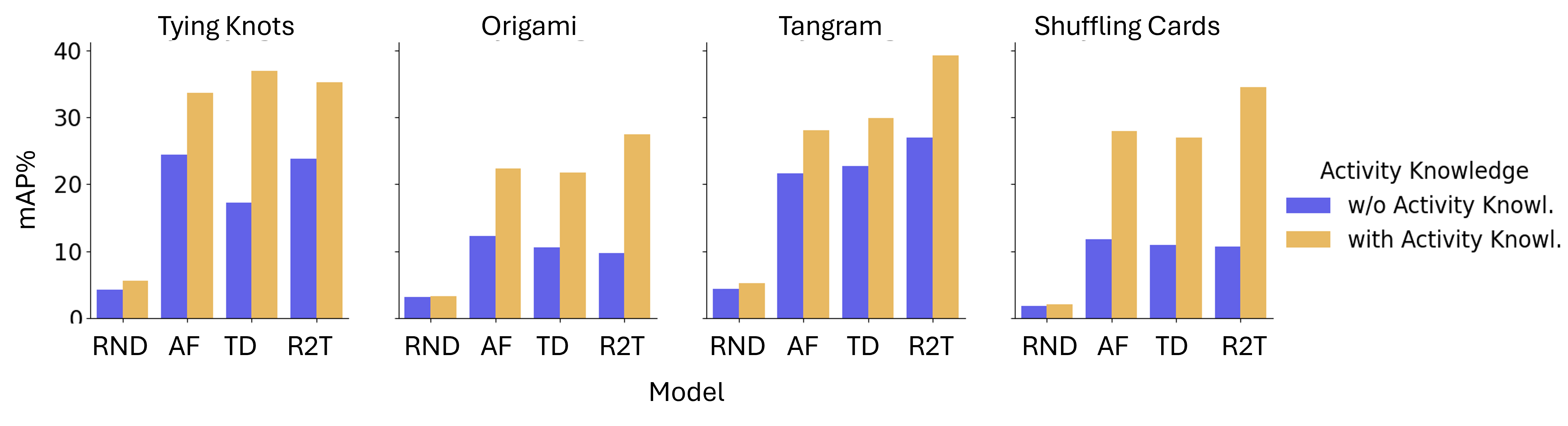}
    \caption{Comparison of models trained with/without Activity knowledge.}
    \label{fig:subactivity-activity-gen-comparison}
\end{subfigure}
\begin{subfigure}{1.\linewidth}
    \centering
    \includegraphics[width=\linewidth]{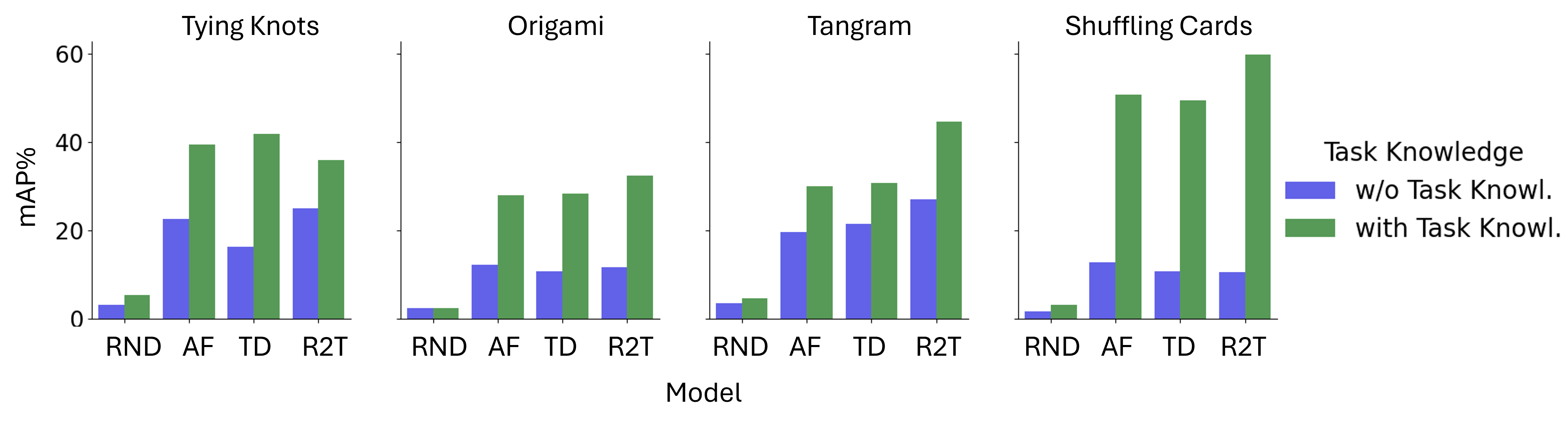}
    \caption{Comparisons of models with/without Task knowledge.}
    \label{fig:activity-within-domain-gen-comparison}
\end{subfigure}
\caption{We showcase the importance of Activity knowledge (a), and Task knowledge in (b). The abbreviations represent the models: RND (Random), AF (Actionformer~\cite{zhang2022actionformer}), TD (TriDet~\cite{10203543}), and R2T (Re2TAL~\cite{Zhao_2023_CVPR}).}
\vspace{-3mm}
\label{fig:combined-generalization-comparison}
\end{figure}

\myparagraph{Importance of Activity/Task Knowledge}
\label{domain comparison}
Here, we wish to evaluate the importance of task-specific and activity-specific knowledge for struggle determination and how this compares across the models tested and the different activities proposed within \datasetname.
Namely, we compare models trained for Activity-Level with those trained for the Task-Level and Within-Activity settings in Fig.~\ref{fig:combined-generalization-comparison}.

Firstly, in Fig.~\ref{fig:subactivity-activity-gen-comparison}, we compare the importance of activity-specific knowledge across the four models.
We note that Tangram has a relatively small drop in performance, indicating that models generalize well to participants struggling with the puzzle and that activity-specific knowledge is less important for this activity.
However, there is a large gap between models with/without shuffling cards knowledge, highlighting its difficulty without activity-specific knowledge.

Secondly, we analyse the importance of task-specific knowledge in Fig.~\ref{fig:activity-within-domain-gen-comparison} by comparing models trained for the Activity Generalization setting with those trained for the Within-Activity setting.
Performance once again drops, with the largest decrease in the Shuffling Cards activity, followed by significant decreases in Origami and Tying Knots without task-specific knowledge.
We note that results are consistent across all models, suggesting an area for future models to exploit.

\subsection{Impact of Skill Evolution on Performance}
\label{struggle-ablation-attempts}

\begin{figure}[ht]
\centering
\includegraphics[width=\linewidth]{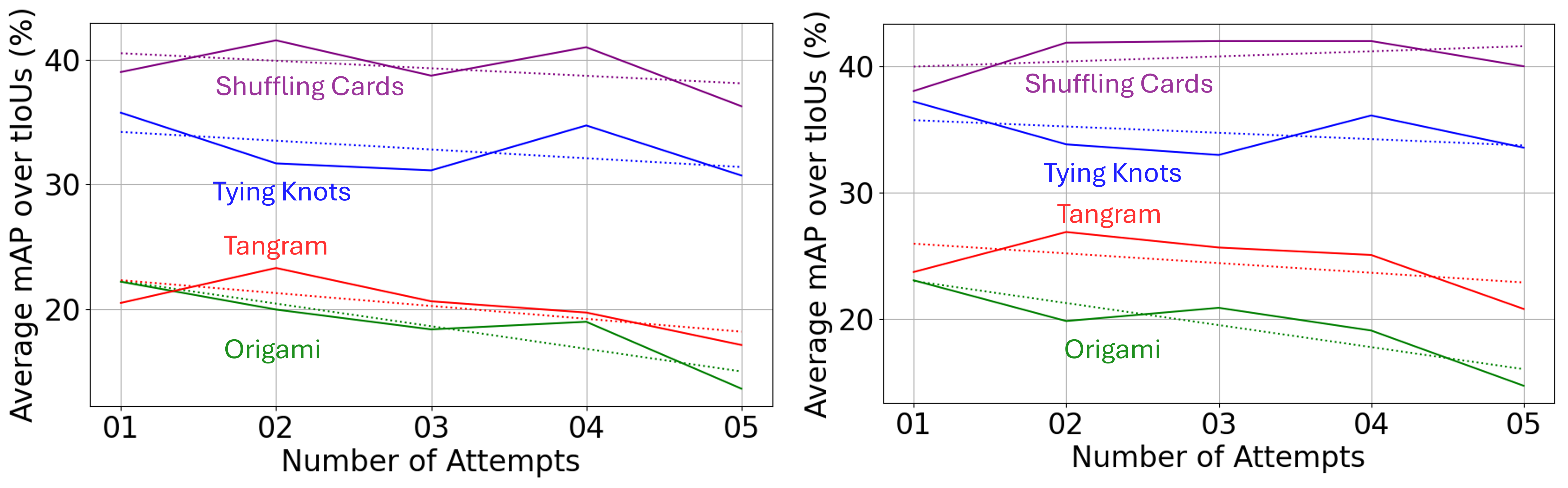}
\caption{Impact of training with individual attempt data. The left line plot presents results for separate attempts using Actionformer~\cite{zhang2022actionformer}, while the right line plot shows results using TriDet~\cite{10203543}. The dashed trend lines highlight the importance of including low-skill participants.}
\vspace{-2mm}
\label{fig:multi-attempts-ablation-study}
\end{figure}

Here, we explore the impact of skill on the temporal struggle localization task. As shown in the statistical bar charts on the right side of Fig.~\ref{fig:combined_struggle_figures}, both struggle time and task completion time decrease as participants repeat the same task, demonstrating their evolution of skill.
This trend raises an important question: How does the deep model's evaluation performance change when trained exclusively on videos from isolated attempts?

To answer this, we ablate models by training using videos from only one attempt.
The models were then evaluated on the validation set for each activity.
Results on Actionformer~\cite{zhang2022actionformer} and TriDet~\cite{10203543} are shown in Fig.~\ref{fig:multi-attempts-ablation-study}.
The trend lines generally show decreasing mAPs as the number of attempts increases.
This decline may be attributed to participants exhibiting fewer instances of struggle as they repeat the same task, resulting in fewer struggle-related data to train the models effectively. This phenomenon also underscores the importance of including low-skill individuals in training datasets to enhance struggle detection performance. Additional experiments on the evolution of skills are shown in supp.

\subsection{Qualitative Analysis of Model Predictions}
\label{struggle-qualitative-analysis}
\begin{figure}[ht]
\begin{center}
\includegraphics[width=1.\linewidth]{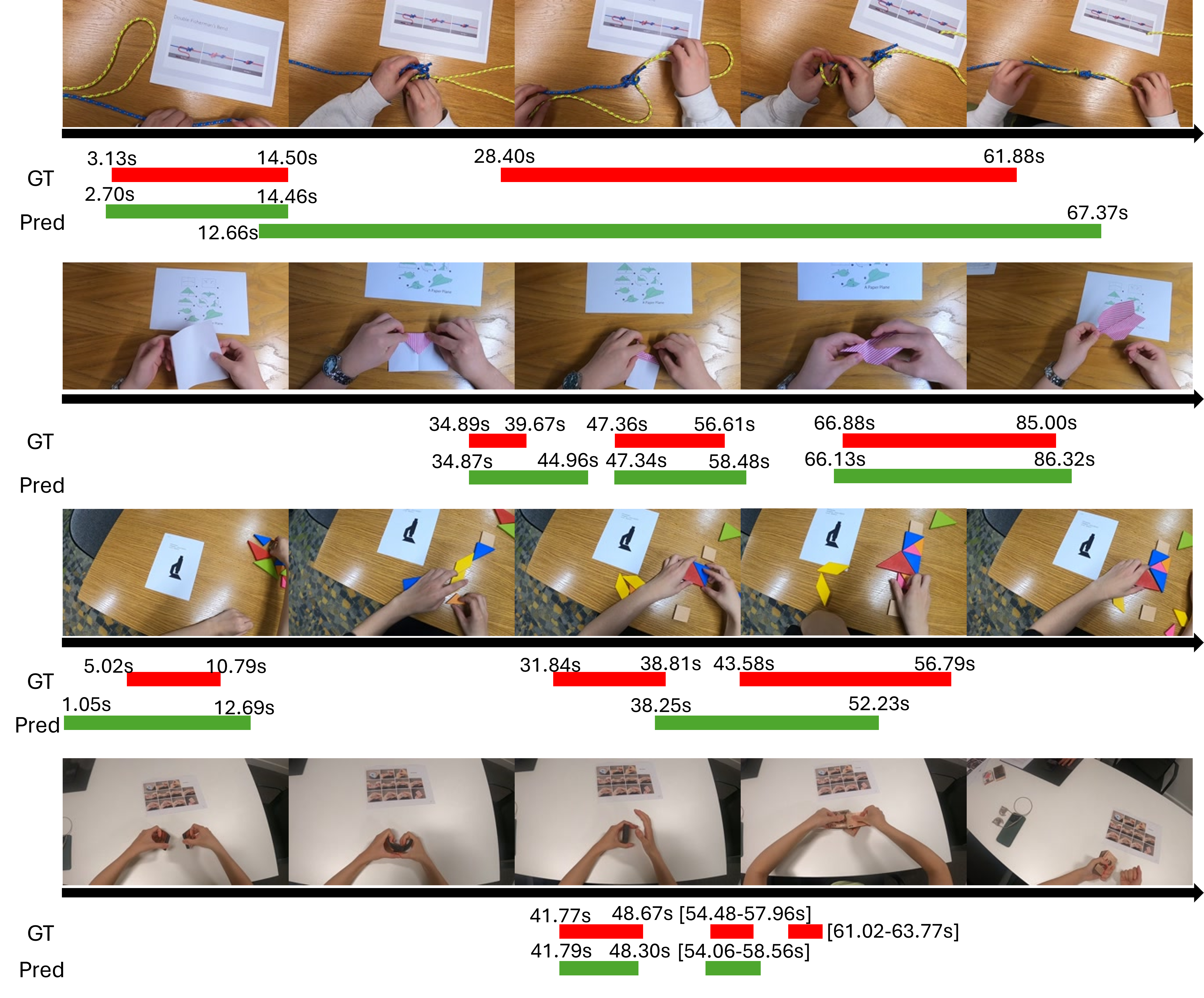}
\end{center}
\vspace{-5mm}
   \caption{Visualizations for the predicted temporal struggle segments vs the ground truth (GT).} 
\label{fig:prediction-gt-visualisation}
\vspace{-5mm}
\end{figure}

In Fig.~\ref{fig:prediction-gt-visualisation}, we visualize predictions of struggle segments using the TriDet model~\cite{10203543} and compare vs the ground truth.
We can observe some misses of small struggle moments, over-prediction of certain segments, or prediction of imprecise struggle boundaries. Although the overall figure shows that TriDet~\cite{10203543} mostly recognizes areas exhibiting struggle, there are gaps.
While temporal struggle action localization is achievable and can yield precise detection results with high IoUs, challenges remain due to the diversity in struggle segment lengths, and the subtle differences between struggle and non-struggle moments. Long struggle segments, which require modelling long-term temporal dependencies, pose a particular challenge. This is evident in the first row (tying knots activity), where the model's predictions do not fully align with the ground truth. Similarly, the struggle moment boundaries are difficult to detect precisely, as seen in the third row (tangram activity). This difficulty likely arises from the subtle differences between struggle moments and normal actions. Short struggle segments can also be challenging to detect, as illustrated in the fourth row (shuffle cards activity), where some struggle instances are missing from the predictions.

\section{Limitations}
\label{limitation}
While our dataset offers valuable insights into struggle detection, some limitations do exist. 
Whilst we chose the activities to capture a broad range of struggling, we cannot cover all real-world tasks or activity types. However, future works can benefit from our generalization experiments, which show that similar tasks/activities may share common patterns for detecting struggle so that more efficient ways to expand struggle data can be considered.

\section{Conclusions}
\label{sec:discussion} 
In this paper, we introduced \datasetname, a large-scale dataset for struggle determination. Our dataset encompasses 61.68 hrs of video with 18 tasks grouped into 4 distinct activities---tying knots, origami, tangram, and shuffling cards.
Each task was repeated five times per participant to capture participants' evolution of their struggle/skill.
We manually annotated struggle segments with start and end times for all videos, creating high-quality annotations for the struggle temporal localization task.
Our experiments highlight the challenge and worth of the dataset across activity/task generalization and evolution of skill.
Results show current models still need progress for high IoU settings, which we hope will encourage future work in this area.

{
    \small
    \bibliographystyle{ieeenat_fullname}
    \bibliography{main}
}

\end{document}


\maketitle

\section{Additional Details on \datasetname{} Dataset}

\subsection{Data Collection}
\label{add-data-collection}

\paragraph{Participant Selection}
To better capture instances of struggle, we prioritized recruiting participants with no prior experience in the activity. A total of 76 participants took part in the dataset collection, including postgraduate and undergraduate students from various departments, five teaching staff, and two individuals from outside the university. All participants were required to be adults (18 years or older) with normal or corrected vision and to provide consent for their data to be used for research purposes.

In accordance with ethical guidelines, data recording focused exclusively on hand-object interactions, ensuring that facial expressions were not captured. This approach protected participant privacy while maintaining the focus on struggle-related actions.

\paragraph{Instructions Provided to Participants}
While video instructions were initially considered, we opted for paper instructions due to several drawbacks of video-based guidance. Although videos can demonstrate tasks in detail, they often require participants to pause and replay, introducing unnecessary hand movements unrelated to the task. Moreover, screen glare during recording could degrade video quality, making paper instructions a more practical choice.

Participants were provided with step-by-step instructions for each activity. In the knot-tying and origami tasks, instructions outlined the key steps necessary to complete the task. In the Tangram puzzle activity, a three-minute time limit was imposed for each puzzle to prevent participants from being stuck for too long. If a participant was unable to complete a puzzle within the time limit, they were stopped and given hints before proceeding with their remaining attempts.

For the card-shuffling activity, participants were instructed to shuffle the cards repeatedly for one to one-and-a-half minutes per attempt. Since card shuffling is typically a quick action, and struggle indicators (e.g., dropping cards) occur briefly, this approach helped capture more struggle instances. Additionally, repeated shuffling reflects natural behaviour, as people often shuffle multiple times per deal in a game.

\subsection{Datasets Statistics}
\label{add-stats}
\begin{figure*}[ht]
\begin{center}
\includegraphics[width=1.0\linewidth]{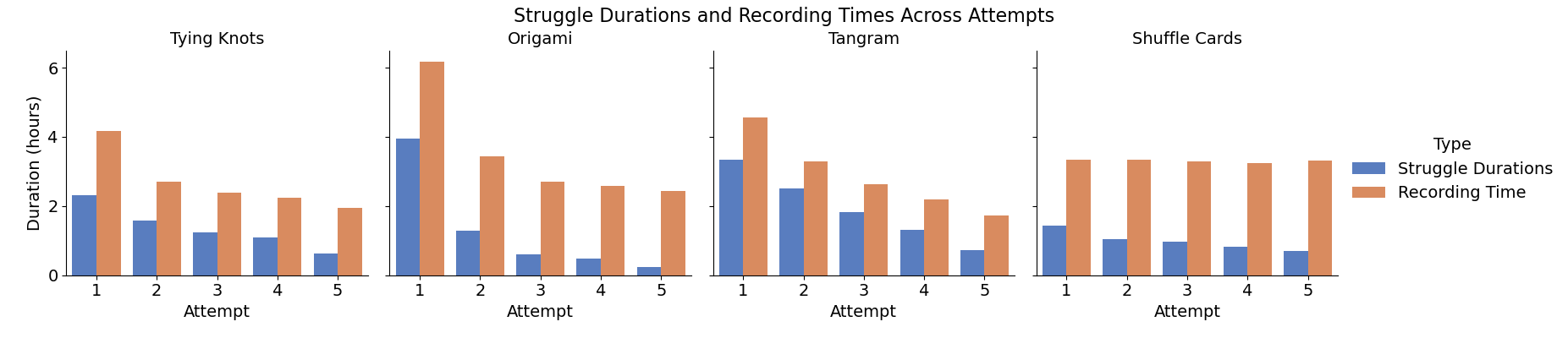}
\end{center}
   \caption{Bar plots showing the summation of video recording times and the struggle durations within each of the five attempts over the four activities.}
\label{fig:struggle_duration_recording_time_across_attempts}
\end{figure*}

\begin{figure*}[ht]
\begin{center}
\includegraphics[width=1.0\linewidth]{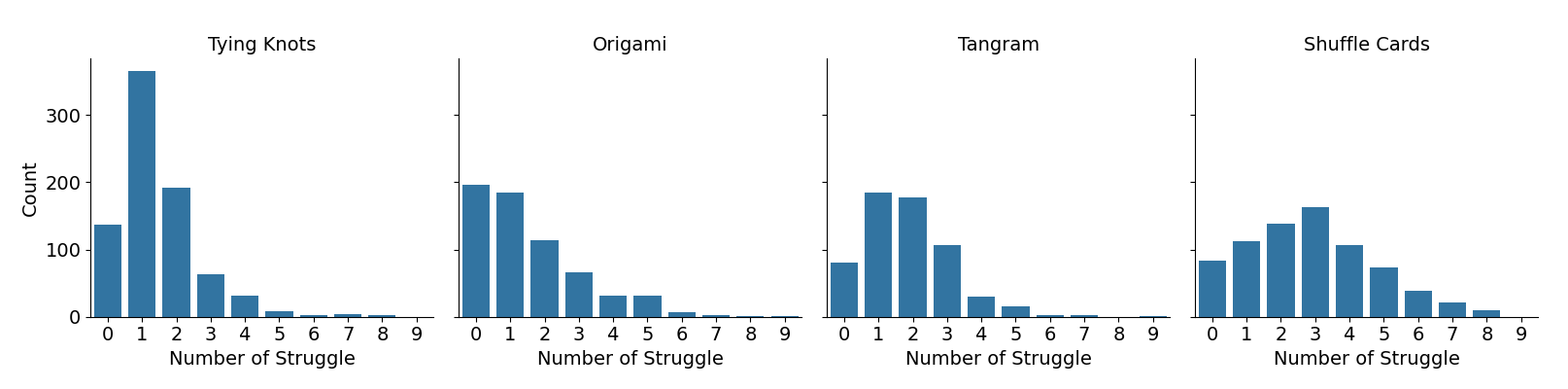}
\end{center}
   \caption{The struggle instances per video for each of the four activities.}
\label{fig:num_struggle_histogram}
\end{figure*}

\begin{figure*}[ht]
\begin{center}
\includegraphics[width=1.0\linewidth]{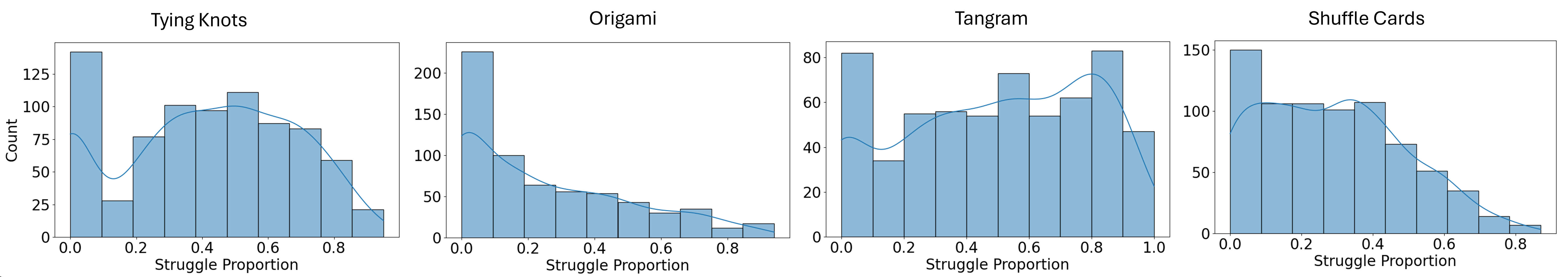}
\end{center}
   \caption{The proportions of struggle duration relative to the video recording duration per video in each of the four activities.}
\label{fig:struggle_proportions}
\end{figure*}

We provide additional statistics for our \datasetname{} dataset. Fig.~\ref{fig:struggle_duration_recording_time_across_attempts} shows the struggle duration and video recording time over the five attempts in the four activities. Both struggle duration and video recording time generally decrease as participants gain more experience in later attempts. However, the recording time for shuffling cards remains similar because we set a fixed recording duration of either one minute or one and a half minutes for each task.
Fig.~\ref{fig:num_struggle_histogram} illustrates the distribution of videos based on the number of struggle instances across the four activities. Videos with 1–2 struggle instances are the most common while shuffling cards generally have a slightly higher number of struggle instances. This is because participants were allowed to shuffle as many times as possible within a given time limit.
Finally, Fig.~\ref{fig:struggle_proportions} presents the proportion of struggle duration relative to the recording time per video, demonstrating the diversity of temporal struggle periods in our dataset.

\subsection{Datasets Comparison}
\label{add-dataset-compare} 
Table~\ref{tab:combined-statistics} presents a statistical comparison between our \datasetname{} and the existing struggle determination dataset~\cite{feng2024strugglingdatasetbaselinesstruggle}. In the existing struggle dataset, videos are divided into 10-second segments, so the reported video count refers to these segments. Struggle instances are estimated by counting segments labelled as struggle, based on expert annotations. Recording duration is computed as the number of segments multiplied by 10 seconds. Per-video struggle instances and struggle duration are calculated by treating each struggle-labelled segment as one instance, with a fixed duration of 10 seconds. Accordingly, total struggle duration (in hours) is estimated as the number of struggle segments multiplied by 10 seconds and then converted to hours.

In contrast, \datasetname{} features untrimmed videos with struggle annotated using precise start and end times. Additionally, each activity includes multiple tasks under the same theme, contributing greater diversity to the struggle determination dataset.

\begin{table*}[t]
\begin{center}
\resizebox{\textwidth}{!}{%
\begin{tabular}{llrrrrrrrr}
\toprule
 & & \multicolumn{5}{c}{Total} & \multicolumn{3}{c}{Per Video} \\ \cmidrule(lr){3-7} \cmidrule(lr){8-10}
Dataset & Activities(\#Tasks) & \#Part. & \#Videos & \#Str. Inst. & Str. Dur. (hrs) & Rec. Dur. (hrs) & \#Str. Inst. & Str. Dur. (s) & Rec. Dur. (s) \\ \midrule
Pipes-Struggle~\cite{feng2024strugglingdatasetbaselinesstruggle} & Pipes (1) & 24 & 1011 & 635 & 1.76 & 2.81 & 0.63$\pm$0.48 & 6.30$\pm$4.80 & 10.00$\pm$0.00 \\
Tent-Struggle~\cite{feng2024strugglingdatasetbaselinesstruggle} & Tent (1) & 29 & 585 & 321 & 0.89 & 1.63 & 0.55$\pm$0.50 & 5.50$\pm$5.00 & 10.00$\pm$0.00 \\
Tower-Struggle~\cite{feng2024strugglingdatasetbaselinesstruggle} & Tower (1) & 20 & 236 & 189 & 0.53 & 0.66 & 0.80$\pm$0.40 & 8.00$\pm$4.00 & 10.00$\pm$0.00 \\ \midrule
EvoStruggle & Tying Knots (5) & 34 & 806 & 1167 & 6.85 & 13.44 & 1.45$\pm$1.20 & 30.61$\pm$36.18 & 60.05$\pm$45.59 \\
EvoStruggle & Origami (4) & 32 & 637 & 974 & 6.54 & 17.32 & 1.53$\pm$1.57 & 36.94$\pm$58.22 & 97.86$\pm$60.79 \\
EvoStruggle & Tangram (4) & 30 & 600 & 1098 & 9.73 & 14.40 & 1.83$\pm$1.30 & 58.38$\pm$59.79 & 86.39$\pm$63.22 \\
EvoStruggle & Shuffle Cards (5) & 30 & 750 & 2146 & 4.98 & 16.52 & 2.86$\pm$1.89 & 23.90$\pm$19.45 & 79.31$\pm$17.66 \\ \midrule
\textbf{EvoStruggle} & \textbf{Total} & \textbf{126 (76)} & \textbf{2793} & \textbf{5385} & \textbf{28.10} & \textbf{61.68} & \textbf{1.93$\pm$1.62} & \textbf{36.22$\pm$46.62} & \textbf{79.50$\pm$50.78} \\ \bottomrule
\end{tabular}
}
\end{center}
\caption{Statistics comparison between the existing struggle determination dataset~\cite{feng2024strugglingdatasetbaselinesstruggle} and our EvoStruggle Datasets. The initial means the struggle determination dataset proposed in~\cite{feng2024strugglingdatasetbaselinesstruggle}. The column ``Activities (\#Tasks)" indicates the activity name and number of tasks within it. Abbreviations: ``\#Parts." for number of participants, ``\#Str. Inst." for number of struggle instances, ``Str. Dur." for struggle duration, ``Rec. Dur." for recording duration. For EvoStruggle (total), ``\#Parts." reflects 126 total participations by 76 unique participants, considering the same person participating in different activities.}
\label{tab:combined-statistics}
\end{table*}

\section{Additional Details of Experiments}
\label{add-experiments}

\subsection{Model Configurations}
The Actionformer model~\cite{zhang2022actionformer} is configured with two different setups based on the video duration of activities: (1) For shorter-duration activities, such as tying knots and shuffling cards, the multi-head self-attention window size is set to 7, with a maximum input feature sequence length of 192; (2) For longer-duration activities, such as origami and tangram, the multi-head self-attention window size is set to 19, with a maximum input feature sequence length of 576. The full sequences of features are input into the model during inference time without using the position embeddings. 

Similarly, for the Scalable-Granularity Perception (SGP) layer in the TriDet~\cite{10203543} model, we configure the window-level branch as follows: for activities such as tying knots and shuffling cards, the temporal depth-wise convolution window size w is set to 3, and the inflating scale factor k is set to 1; for activities such as origami and tangram, we increase the inflating scale factor to 3 while keeping the window size remains the same. We adopt the same configurations of the maximum input sequence lengths in the training stage and also use the full sequence input at the inference time. 

The end-to-end TAL model Re2TAL~\cite{Zhao_2023_CVPR} was implemented with a rewired reversible version of SlowFast-101~\cite{feichtenhofer2019slowfastnetworksvideorecognition} as the feature extractor, followed with Actionformer~\cite{zhang2022actionformer} as the temporal action detection head. We load the pre-trained parameters of the reversible SlowFast-R101~\cite{feichtenhofer2019slowfastnetworksvideorecognition} on Kinetics-400~\cite{https://doi.org/10.48550/arxiv.1705.06950} before training starts. The action detection head has a fixed input feature sequence length of 192 for all activities. The input frame sequence length is uniformly resized to 382 frames for training and inference time. This limitation is due to GPU memory constraints when using long sequences of video frames as input.

\subsection{Implementation Details}
We uniformly resize the video frame resolution from 1080p to 360p, and the frame rates remain the same using FFmpeg so that it is possible to load the video into the bowser-based video annotation software, VIA Video Annotator~\cite{dutta2019vgg}. Loading the video data during the end-to-end training process also takes less time.

We extract features for the feature-based TAL models using SlowFast-R50~\cite{feichtenhofer2019slowfastnetworksvideorecognition} pre-trained on Kinetics-400~\cite{https://doi.org/10.48550/arxiv.1705.06950} using clips of 32 frames at a frame rate of 50 fps and a stride of 16 frames. This gives one feature vector per 0.32 seconds. The features are extracted at 3.125 features per second with dimension 2304.

We add Gaussian noise with $\sigma=0.05$ into the input features when training the Actionformer~\cite{zhang2022actionformer} to alleviate the overfitting. And $\sigma=0.0005$ for the model TriDet~\cite{10203543}. For the end-to-end training using the model Re2TAL~\cite{Zhao_2023_CVPR}, the video frames are loaded from a size of 360p and firstly resized into $256\times256$ and then randomly cropped into $224\times224$ during the training stage, with the horizontal flip at 0.5 probability and colour jittering. At the same time, we use the centre crop without flipping and colour jittering in the evaluation stage, which includes the validation and test. 

In the task-level generalization experiments, we set the base learning rate to be 1e-4 using the cosine scheduler, and the models are optimized using Adam optimizer and trained for 35 epochs, with the first 5 epochs being the linear warm-up stages, for both Actionformer~\cite{zhang2022actionformer} and TriDet~\cite{10203543}. The weight decay is 0.05 for Actionformer~\cite{zhang2022actionformer} and 0.025 for TriDet~\cite{10203543}. The training batch size is 8, and the test batch size is one for the full sequence input. The maximum number of segmentation predicted is set to be 10 throughout all our experiments. For the Re2TAL-Slowfast101-Actionformer~\cite{Zhao_2023_CVPR}, we set the base learning rate to be 1e-3, and the base learning rate for the backbone is 1e-5. We trained for 25 epochs with the cosine learning rate scheduler and ran the first 3 epochs as the linear warm-up stage. The batch size is limited to 2 with two 1080 Ti GPUs. 

In the activity-level generalization experiments, the hyperparameter settings are mostly similar to those in the task-level generalization experiments. However, we increase the training epochs to 50, with a larger batch size of 16 for the Actionformer~\cite{zhang2022actionformer} and TriDet~\cite{10203543}, 8 for the model Re2TAL~\cite{Zhao_2023_CVPR} trained end-to-end using one A100 GPU, because of the more training data. We adopt similar hyperparameter settings in our within-activity evaluation baseline experiments. 

\subsection{Full Details of Experiment Results}

We present the full experimental results, reporting mAP values across a full range of IoU thresholds from 0.3 to 0.7. Table~\ref{tab:within-domain-eval-fulltab} shows the results for within-activity evaluation, and Table~\ref{tab:activity-level gen fulltab} provides the results for activity-level generalization experiments.

\begin{table*}[ht]
\scriptsize
\centering
\begin{tabular}{ll|cccccc}
\toprule
 &  & \multicolumn{6}{c}{Held-out Test mAP@tIoU[0.3:0.1:0.7] and Average} \\ \cmidrule{3-8}
Activity & Model & 0.3 & 0.4 & 0.5 & 0.6 & 0.7 & Avg. \\ \midrule
\multirow{4}{*}{Tying Knots} & Random & 8.80\% & 4.50\% & 1.79\% & 0.58\% & 0.19\% & 3.17\% \\
 & Actionformer~\cite{zhang2022actionformer} & 67.99\% & 54.56\% & 39.21\% & 24.79\% & 10.38\% & 39.39\% \\
 & TriDet~\cite{10203543} & 65.44\% & 56.60\% & 43.91\% & 29.15\% & 14.53\% & \textbf{41.93\%} \\
 & Re2TAL~\cite{Zhao_2023_CVPR} & 61.73\% & 51.01\% & 38.12\% & 20.04\% & 8.70\% & 35.92\% \\ \midrule
\multirow{4}{*}{Origami} & Random & 7.25\% & 3.61\% & 0.97\% & 0.19\% & 0.06\% & 2.42\% \\
 & Actionformer~\cite{zhang2022actionformer} & 52.75\% & 39.12\% & 27.73\% & 15.05\% & 5.23\% & 27.98\% \\
 & TriDet~\cite{10203543} & 54.32\% & 41.40\% & 27.00\% & 13.43\% & 5.73\% & 28.38\% \\
 & Re2TAL~\cite{Zhao_2023_CVPR} & 57.37\% & 46.44\% & 29.62\% & 19.59\% & 8.78\% & \textbf{32.36\%} \\ \midrule
\multirow{4}{*}{Tangram} & Random & 10.29\% & 4.28\% & 1.90\% & 0.64\% & 0.20\% & 3.46\% \\
 & Actionformer~\cite{zhang2022actionformer} & 55.85\% & 43.44\% & 30.64\% & 14.94\% & 4.90\% & 29.95\% \\
 & TriDet~\cite{10203543} & 57.21\% & 44.97\% & 29.77\% & 15.33\% & 6.21\% & 30.70\% \\
 & Re2TAL~\cite{Zhao_2023_CVPR} & 69.27\% & 58.11\% & 44.83\% & 31.54\% & 19.11\% & \textbf{44.57\%} \\ \midrule
\multirow{4}{*}{Shuffle Cards} & Random & 5.07\% & 2.03\% & 0.84\% & 0.28\% & 0.08\% & 1.66\% \\
 & Actionformer~\cite{zhang2022actionformer} & 71.49\% & 63.12\% & 56.40\% & 41.12\% & 21.85\% & 50.80\% \\
 & TriDet~\cite{10203543} & 70.94\% & 61.66\% & 55.26\% & 39.60\% & 20.28\% & 49.55\% \\
 & Re2TAL~\cite{Zhao_2023_CVPR} & 78.26\% & 70.07\% & 62.30\% & 53.01\% & 35.21\% & \textbf{59.77\%} \\ \bottomrule \\
\end{tabular}%
\caption{Within-Activity Evaluation Experiment Results. The results are reported on the validation set in each activity.}
\label{tab:within-domain-eval-fulltab}
\end{table*}

\begin{table*}[ht]
\scriptsize
\centering
\begin{tabular}{ll|cccccc}
\toprule
 &  & \multicolumn{6}{c}{Held-out Test mAP@tIoU[0.3:0.1:0.7] and Average} \\ \cmidrule{3-8}
Activity & Model & 0.3 & 0.4 & 0.5 & 0.6 & 0.7 & Avg. \\ \midrule
\multirow{4}{*}{Tying Knots} & Random & 8.80\% & 4.50\% & 1.79\% & 0.58\% & 0.19\% & 3.17\% \\
 & Actionformer~\cite{zhang2022actionformer} & 45.13\% & 30.03\% & 20.37\% & 13.06\% & 4.30\% & 22.58\% \\
 & TriDet~\cite{10203543} & 34.76\% & 23.65\% & 14.08\% & 6.07\% & 2.68\% & 16.25\% \\
 & Re2TAL~\cite{Zhao_2023_CVPR} & 47.14\% & 35.87\% & 23.45\% & 13.61\% & 5.19\% & \textbf{25.05\%}\\ \midrule
\multirow{4}{*}{Origami} & Random & 7.25\% & 3.61\% & 0.97\% & 0.19\% & 0.06\% & 2.42\% \\
 & Actionformer~\cite{zhang2022actionformer} & 32.07\% & 15.70\% & 8.54\% & 3.92\% & 0.99\% & \textbf{12.24\%} \\
 & TriDet~\cite{10203543} & 28.78\% & 13.53\% & 7.08\% & 3.11\% & 1.12\% & 10.72\% \\
 & Re2TAL~\cite{Zhao_2023_CVPR} & 26.26\% & 15.66\% & 9.14\% & 4.65\% & 2.65\% & 11.67\%\\ \midrule
\multirow{4}{*}{Tangram} & Random & 10.29\% & 4.28\% & 1.90\% & 0.64\% & 0.20\% & 3.46\% \\
 & Actionformer~\cite{zhang2022actionformer} & 44.58\% & 28.21\% & 15.60\% & 6.79\% & 2.83\% & 19.60\% \\
 & TriDet~\cite{10203543} & 47.07\% & 30.40\% & 18.19\% & 8.20\% & 3.23\% & 21.42\% \\
 & Re2TAL~\cite{Zhao_2023_CVPR} & 49.53\% & 36.79\% & 24.12\% & 15.49\% & 8.97\% & \textbf{26.98\%} \\ \midrule
\multirow{4}{*}{Shuffle Cards} & Random & 5.07\% & 2.03\% & 0.84\% & 0.28\% & 0.08\% & 1.66\% \\
 & Actionformer~\cite{zhang2022actionformer} & 29.15\% & 18.11\% & 9.86\% & 4.92\% & 1.40\% & \textbf{12.69\%} \\
 & TriDet~\cite{10203543} & 28.42\% & 14.75\% & 7.15\% & 2.74\% & 0.67\% & 10.75\% \\
 & Re2TAL~\cite{Zhao_2023_CVPR} & 24.80\% & 15.20\% & 8.00\% & 3.65\% & 0.98\% & 10.53\% \\ \bottomrule \\
\end{tabular}%
\caption{Activity-Level Generalization Experiment Results. The results are reported based on the validation set in each activity as the held-out test activity.}
\label{tab:activity-level gen fulltab}
\end{table*}

\subsection{Training on Sampled Subsets}
In the experiments investigating the impact of skill evolution, we created three sampled sets for each activity, each consisting of five subsets. Each subset contained one-fifth of the videos from individual attempts, ensuring that the number of videos in each final sampled set matched the number in a single attempt for a fair comparison. This sampling process was repeated for each activity.

Deep models were then trained separately on each subset and evaluated on the same validation set. The experiment results, shown in Table~\ref{tab:train_sampled_subsets}, indicate that including all five attempts does not significantly improve overall struggle detection performance when maintaining the same training set size for a fair comparison. The performance remains at an average level compared to training on individual attempts.

\subsection{Training and Evaluation Across Attempts}

To further investigate the dependencies of the struggle corresponding to different attempts in skill evolution, we trained five separate Actionformer~\cite{zhang2022actionformer} models, each using data from a specific attempt (i.e., the 1st to 5th repetitions). Each model was then evaluated on all five attempts, resulting in a 5×5 performance matrix that captures the cross-repetitions generalization. This setup allows us to examine potential dependencies between repetitions.

\begin{figure*}[ht]
\begin{center}
\includegraphics[width=1.0\linewidth]{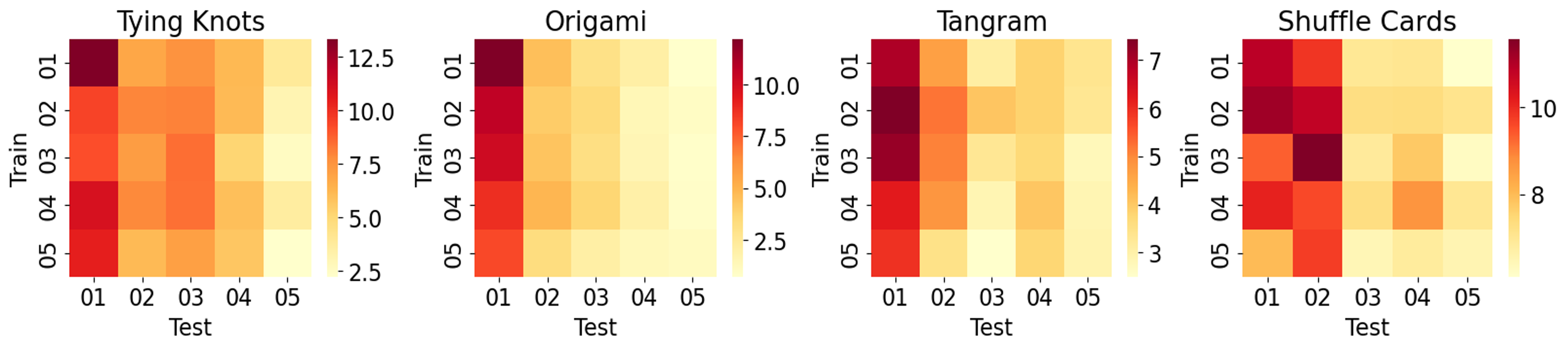}
\end{center}
   \caption{Heatmaps showing the model Actionformer~\cite{zhang2022actionformer} trained on each attempt and evaluated on each attempt within the validation split.}
\label{fig:further-experiment-results-on-multiple-attempts}
\end{figure*}

As shown in Fig.~\ref{fig:further-experiment-results-on-multiple-attempts}, models consistently achieved the highest mean Average Precision (mAP) when evaluated on the first repetition, regardless of the training repetition. In contrast, performance was lowest when evaluated on the fifth repetition. These findings suggest that struggles occurring during early attempts are more pronounced and thus easier to detect, while later repetitions reflect more refined performance with subtler struggle cues.

\begin{table*}[ht]
    \centering
    \scriptsize
    \begin{tabular}{l l c c c c c c}
        \toprule
        Activity & Model & Attempt 1 & Attempt 2 & Attempt 3 & Attempt 4 & Attempt 5 & All Attempts (Subsets) \\
        \midrule
        \multirow{2}{*}{Tying Knots} & Actionformer & 35.75\% & 31.69\% & 31.13\% & 34.73\% & 30.72\% & 30.79\% \\
                                     & TriDet      & 37.22\% & 33.83\% & 32.99\% & 36.12\% & 33.57\% & 32.85\% \\
        \midrule
        \multirow{2}{*}{Origami} & Actionformer & 22.21\% & 19.98\% & 18.39\% & 19.00\% & 13.64\% & 21.08\% \\
                                  & TriDet      & 23.05\% & 19.82\% & 20.87\% & 19.07\% & 14.69\% & 20.60\% \\
        \midrule
        \multirow{2}{*}{Tangram} & Actionformer & 20.51\% & 23.31\% & 20.64\% & 19.74\% & 17.14\% & 20.96\% \\
                                  & TriDet      & 23.71\% & 26.88\% & 25.65\% & 25.06\% & 20.78\% & 24.00\% \\
        \midrule
        \multirow{2}{*}{Shuffle Cards} & Actionformer & 39.01\% & 41.55\% & 38.72\% & 41.00\% & 36.27\% & 40.16\% \\
                                        & TriDet      & 38.06\% & 41.89\% & 42.02\% & 42.02\% & 40.02\% & 40.22\% \\
        \bottomrule
    \end{tabular}
    \caption{Performance comparison between models trained separately on each of the five attempts and the average performance of models trained on sampled subsets containing all attempts.}
    \label{tab:train_sampled_subsets}
\end{table*}

{
    \small
    \bibliographystyle{ieeenat_fullname}
    \bibliography{main}
}